\documentclass[11pt,a4paper]{article}
\usepackage[hyperref]{acl2021}
\usepackage{times}
\usepackage{latexsym}

\usepackage{microtype}

\aclfinalcopy 

\usepackage{soul}
\usepackage[utf8]{inputenc}
\usepackage{amsmath}
\usepackage{amsthm}
\usepackage{booktabs}
\usepackage{algorithm}
\usepackage{subfigure}
\usepackage{multirow}
\usepackage{enumitem}
\usepackage{bm}
\usepackage{amsfonts}
\usepackage{colortbl}
\usepackage[switch]{lineno}
\usepackage{graphicx}
\usepackage{framed}

\usepackage{algpseudocode}
\floatname{algorithm}{Algorithm}

\newcommand{\ie}{\emph{i.e., }}
\newcommand{\eg}{\emph{e.g., }}

\newcommand{\cf}{\emph{cf. }}

\title{Empowering Language Understanding with Counterfactual Reasoning}

\author{Fuli Feng\textsuperscript{1,2},~Jizhi Zhang\textsuperscript{3},~Xiangnan He\textsuperscript{3}$\thanks{$^{*}$Corresponding author.}$,~Hanwang Zhang\textsuperscript{4},~\textbf{Tat-Seng Chua}\textsuperscript{2}
\\
\textsuperscript{1}NExT-Sea Joint Lab,~\textsuperscript{2}National University of Singapore\\\textsuperscript{3}University of Science and Technology of China,~\textsuperscript{4}Nanyang Technological University\\

\tt{fulifeng93@gmail.com},~\tt {cdzhangjizhi@mail.ustc.edu.cn}\\\tt{xiangnanhe@gmail.com},\tt{hanwangzhang@ntu.edu.sg},\tt{dcscts@nus.edu.sg}}

\date{}

\begin{document}
\maketitle
\begin{abstract}
Present language understanding methods have demonstrated extraordinary ability of recognizing patterns in texts via machine learning. However, existing methods indiscriminately use the recognized patterns in the testing phase that is inherently different from us humans who have \textit{counterfactual thinking}, \eg to scrutinize for the hard testing samples. Inspired by this, we propose a \textit{Counterfactual Reasoning Model}, which mimics the counterfactual thinking by learning from few counterfactual samples. In particular, we devise a \textit{generation module} to generate representative counterfactual samples for each factual sample, and a \textit{retrospective module} to retrospect the model prediction by comparing the counterfactual and factual samples. Extensive experiments on sentiment analysis (SA) and natural language inference (NLI) validate the effectiveness of our method.
\end{abstract}

\section{Introduction}
Language understanding~\cite{ke2020sentilare} is a central theme of artificial intelligence~\cite{chomsky2002syntactic}, which empowers a wide spectral of applications such as sentiment evaluation~\cite{feldman2013techniques}, commonsense inference~\cite{snliemnlp2015}. The models are trained on labeled data to recognize the textual patterns closely correlated to different labels. Owing to the extraordinary representational capacity of deep neural networks, the models can well recognize the pattern and make prediction accordingly~\cite{jacob2019bert}. 
However, the cognitive ability of these data-driven models is still far from human beings due to lacking \textit{counterfactual thinking}~\cite{pearl2019seven}.

Counterfactual thinking is a high-level cognitive ability beyond pattern recognition~\cite{pearl2019seven}. 
In addition to observing the patterns within factual samples, counterfactual thinking calls for comparing the fact with imaginations, so as to make better decision.
For instance, 
given a factual sample ``\textit{What do lawyers do when they die? Lie still.}'', the intuitive evaluation of its sentiment based on the textual patterns will recognize ``\textit{Lie still}'' as an objective description of body posture which is \textit{neutral}. By scrutinizing that the ``still'' could be intentionally postposed, we can imagine a counterfactual sample ``\textit{What do lawyers do when they die? \underline{Still lie}.}'' and uncover the negative sarcastic pun, whose sentiment is more accurate.

Recent work~\cite{kaushik2019cfdata,zeng2020counterfactual} shows that incorporating counterfactual samples into model training improves the generalization ability. However, these methods follow the standard machine learning paradigm that uses the same procedure (\eg a forward propagation) to make prediction in the testing phase. That is, making decision for testing samples according to their relative positions to the model decision boundary. The indiscriminate procedure focuses on the textual patterns occurred in the testing sample and treats all testing samples equally, which easily fails on hard samples (\cf Figure~\ref{fig:conf-acc}). 
On the contrary, humans can discriminate hard samples and ponder the decision with a rational system~\cite{daniel2017thinking}, which imagines counterfactual and adjusts the decision.

The key to bridge this gap lies in imitating the counterfactual thinking ability of humans, \ie learning a decision making procedure to serve for the testing phase. That is a procedure of: 1) constructing counterfactual samples for a target factual sample; 2) calling the trained language understanding model to make prediction for the counterfactual samples; and 3) comparing the counterfactual and factual samples to retrospect the model prediction.  
However, the procedure is non-trivial to achieve for two reasons: 1) the space of counterfactual sample is huge since any variant from the target factual sample can be a counterfactual sample. It is thus challenging to search for suitable counterfactual samples that can facilitate the decision making. 2) The mechanism of how we retrospect the decision is still unclear, making it hard to be imitated.

Towards the target, we propose a \textit{Counterfactual Reasoning Model} (CRM), which is a two-phase procedure consisting a \textit{generation module} and a \textit{retrospection module}. In particular, given a factual sample in the testing phase, the generation module constructs representative counterfactual samples by imagining \textit{what would the content be if the label of the sample is $y$}. To imitate the unknown retrospection mechanism of humans, we build the retrospection module as a carefully designed deep neural network that separately compares the latent representation and the prediction of the factual and counterfactual samples. The proposed CRM forms a general paradigm that can be applied to most existing language understanding models without constraint on the format of the language understanding task. We select two language understanding tasks: SA and NLI, and test CRM on three representative models for each task. Extensive experiments on benchmark datasets validate the effectiveness of CRM, which achieves performance gains ranging from 5.1\% to 15.6\%.

The main contributions are as follow:
\begin{itemize}[leftmargin=*]
    \item We propose the \textit{Counterfactual Reasoning Model} to enlighten the language understanding model with counterfactual thinking.
    \item We devise a generation module and a retrospection module that are task and model agnostic.
    \item We conduct extensive experiments, which validate the rationality and effectiveness of the proposed method.
\end{itemize}
\section{Pilot Study}\label{sec:pilot}
Decisions are usually accompanied by confidence, a feeling of being wrong or right~\cite{boldt2019confidence}. From the perspective of model confidence, we investigate the performance of language understanding models across different testing samples. We estimate the model confidence on a sample as the widely used \textit{Maximum Class Probability} (MCP)~\cite{corbiere2019addressing}, which is the probability over the predicted class. A lower value of MCP means less confidence and ``hard'' sample. According to the value of MCP, we rank the testing samples in ascending order and split them into ten groups, \ie confidence level from 1 to 10.

\begin{figure}[tbp]
	\centering
	\mbox{
	    \hspace{-0.15in}
		\subfigure[Sentiment analysis]{
			\label{fig:conf-acc-sa}
			\includegraphics[width=0.24\textwidth]{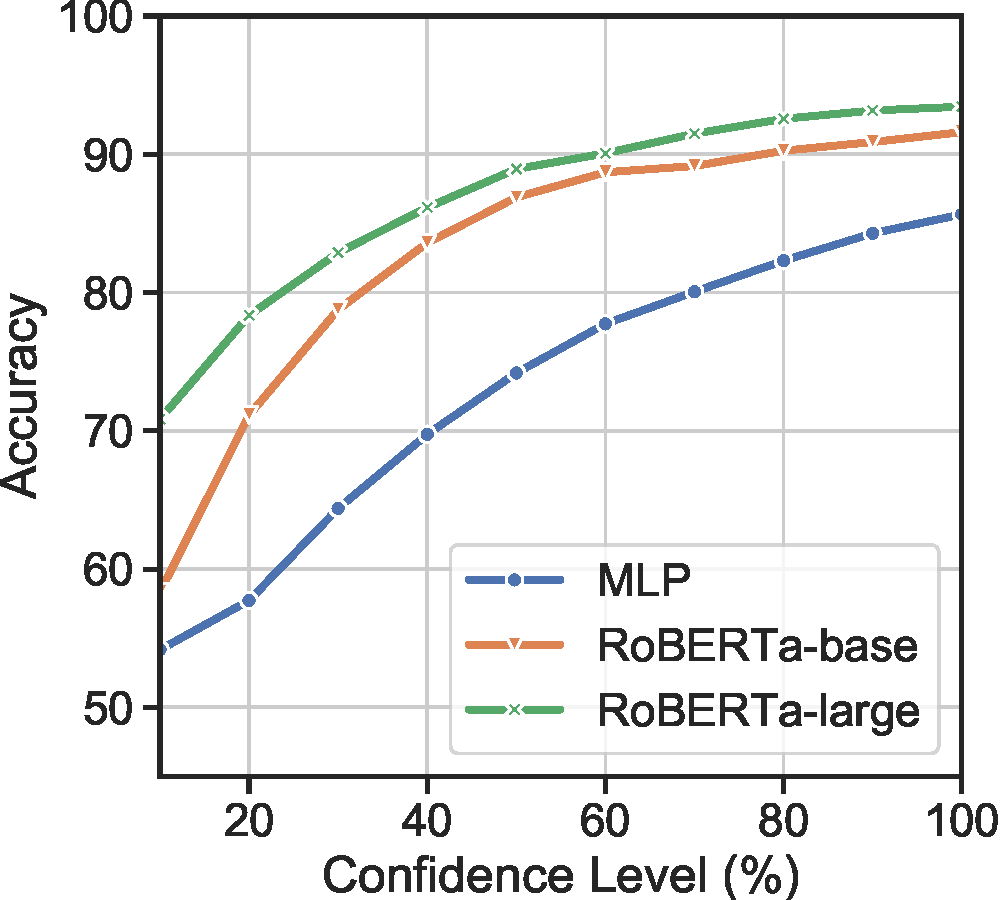}
		}
		\hspace{-0.15in}
		\subfigure[Natural language inference]{
			\label{fig:conf-acc-nli}
			\includegraphics[width=0.24\textwidth]{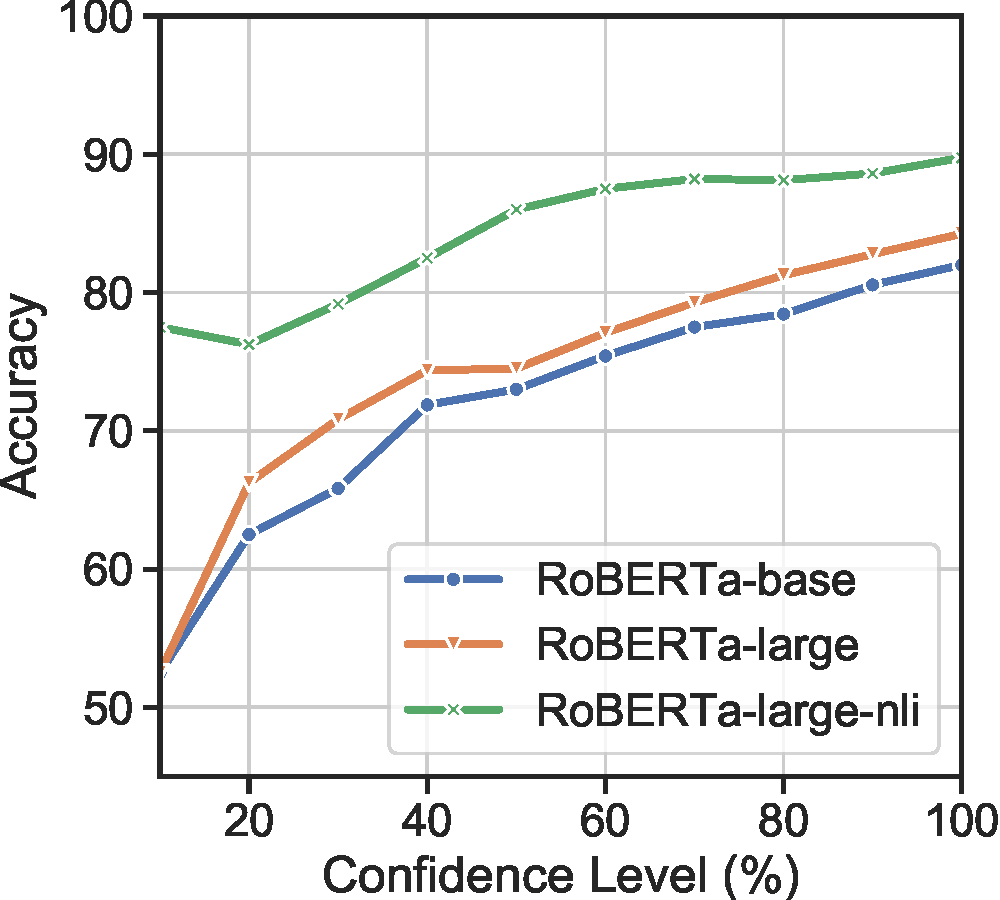}
		}
	}
	\vspace{-0.4cm}
	\caption{Prediction performance of the language understanding models over testing samples at different confidence levels.
	}
	\vspace{-0.3cm}
	\label{fig:conf-acc}
\end{figure}

Figure~\ref{fig:conf-acc} shows the performance of representative models over samples at different model confidence levels on the SA and NLI tasks (see Section~\ref{ssec:setting} for model and dataset descriptions). From the figures, we can observe a clear increasing trend of classification accuracy as the confidence level increases from 1 to 10 in all cases. In other words, these models fail to predict accurately for the hard samples. It is thus essential to enhance the standard inference with a more precise decision making procedure.

\section{Methodology}
In this section, we first formulate the task of learning a decision making procedure for the testing phase (Section~\ref{ssec:formulation}), followed by introducing the proposed CRM (Section~\ref{ssec:crp}) and the paradigm of building language understanding solutions with CRM (Section~\ref{ssec:paradigm}).

\subsection{Problem Formulation}\label{ssec:formulation}
As discussed in the previous work~\cite{wu2020corefqa,li2020unified,li2019entity}, language understanding tasks can be abstracted as a classification problem where the input is a text and the target is to make decision across a set of candidates of interests. 
We follow the problem setting with consideration of counterfactual samples~\cite{kaushik2019cfdata,liang2020col}, where the training data are twofold: 1) \textit{factual samples} $\mathcal{T} = \left\{({\bm{x}}, {{y}})\right\}$ where ${{y}} \in [1, C]$ denotes the class or the target decision of the text; ${\bm{x}} \in \mathbb{R}^{D}$ is the latent representation of the text, which encodes the textual contents\footnote{The input is indeed the plain text which is projected to a latent representation by an encoder (\eg a Transformer~\cite{jacob2019bert}) in the cutting edge solutions. We omit the encoder for briefness since focusing on the decision making.}. 
2) \textit{counterfactual samples} $\mathcal{T}^* = \left\{({\bm{x}}^*_c, c) | ({\bm{x}}, {{y}}) \in \mathcal{T}, c \in [1, C]\&c \neq {{y}} \right\}$ where $({\bm{x}}^*_c, c)$ is a counterfactual sample in class $c$ corresponds to the factual sample $({\bm{x}}, {{y}})$\footnote{Given the labeled factual sample, counterfactual samples can be constructed either manually~\cite{kaushik2019cfdata} or automatically~\cite{chen2020counterfactual} by conducting minimum changes on ${\bm{x}}$ to swap its label from ${{y}}$ to $c$}. 
We assume that a classification model (\eg BERT~\cite{jacob2019bert}) has been trained over the labeled data. Formally,
\begin{equation}
{\small
\begin{aligned}\label{eq:obj-model}
    \hat{\bm{\theta}} = \min_{\bm{\theta}} \sum_{\left({\bm{x}}, {{y}}\right) \in \mathcal{T} / \mathcal{T}^*} l({{y}}, f({\bm{x}} | \bm{\theta})) + \alpha \|\bm{\theta}\|,
\end{aligned}
}%
\end{equation}
where $\hat{\bm{\theta}}$ is the learned parameters of the model $f(\cdot)$
; $l(\cdot)$ is a classification loss such as cross-entropy~\cite{kullback1997information}, and $\alpha$ is a hyper-parameter to adjust the regularization. 

The target is to build a decision making procedure to perform counterfactual reasoning when serving for the testing phase. Given a testing sample $\bm{x}$, the core is a policy of generating counterfactual samples and retrospecting the decision, which is formulated as:
\begin{equation}
{\small
\begin{aligned}
    \bm{y} = 
    h\left(
        \bm{x}, \{\bm{x}^*\}
        | {\bm{\eta}}, \hat{\bm{\theta}}
    \right),~
    \{\bm{x}^*\} = 
    g\left(
        \bm{x} \big| {\bm{\omega}}
    \right), \notag
\end{aligned}
}%
\end{equation}
$\bm{y} \in \mathbb{R}^C$ denotes the final prediction for the testing sample $\bm{x}$, which is a distribution over the classes; $\bm{x}^*$ is one of the generated counterfactual samples for $\bm{x}$. The generation module $g(\cdot)$ parameterized by $\bm{\omega}$ is expected to construct a set of representative counterfactual samples for the target factual sample, which provide signals for the retrospection module $h(\cdot)$ parameterized by $\bm{\eta}$ to retrospect the prediction $f\big(\bm{x} | \hat{\bm{\theta}}\big)$ given by the trained classification model. In particular, $h(\cdot)$ and $g(\cdot)$ will be learned from the factual and counterfactual training samples, respectively.

\subsection{Counterfactual Reasoning Model}\label{ssec:crp}
\begin{figure}[t]
	\centering
	\includegraphics[width=0.45\textwidth]{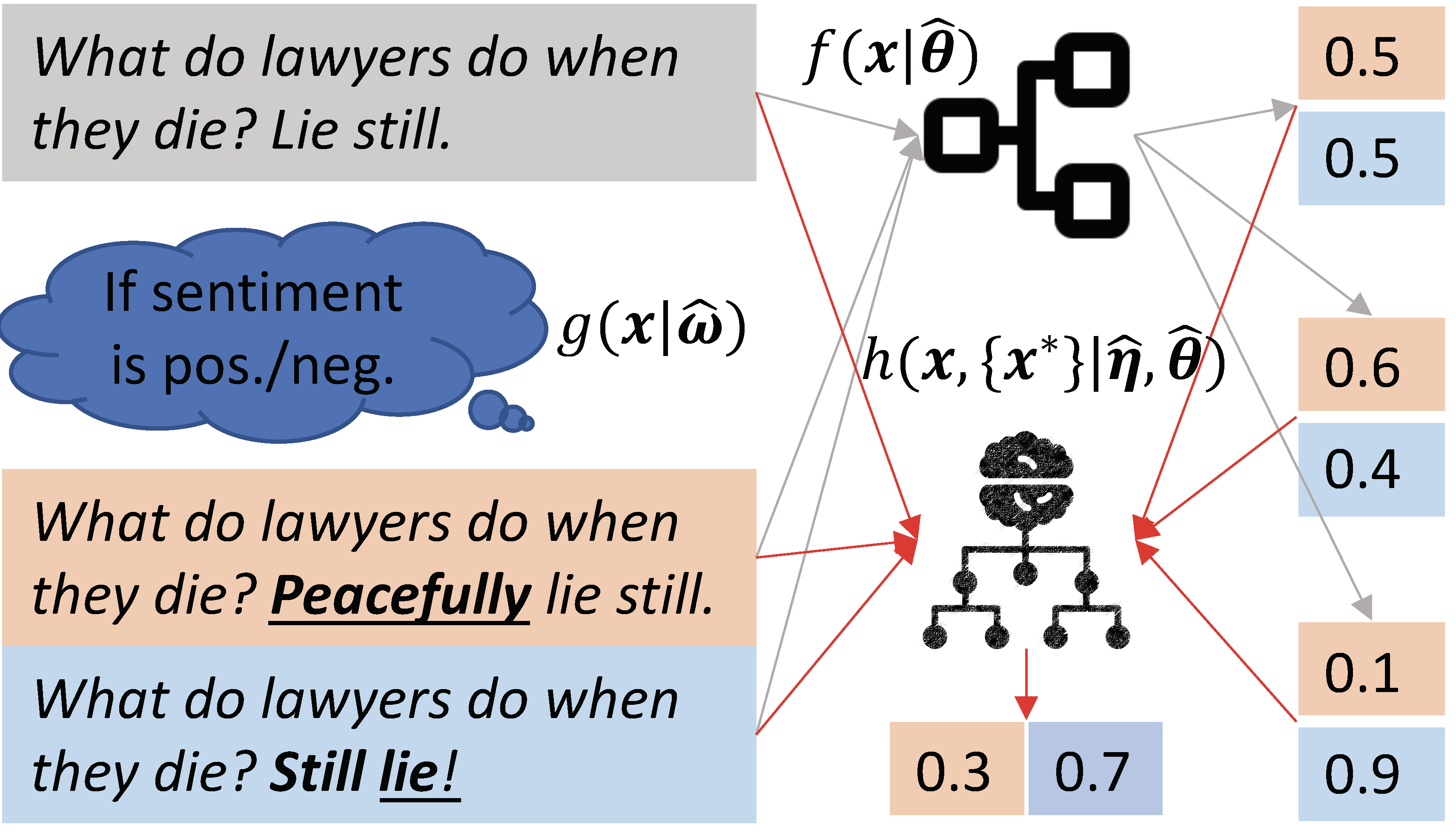}
	\vspace{-0.3cm}
	\caption{Illustration of the proposed CRM.
	}
	\vspace{-0.5cm}
	\label{fig:framework}
\end{figure}

Figure~\ref{fig:framework} illustrates the process of CRM where the arrows in grey color represent the standard inference of trained classification model, and arrows in red color represent the retrospection with consideration of counterfactual samples.

\subsubsection{Retrospection Module}
We devise the retrospection module with one key consideration---distilling signals for making final decision by comparing both the latent representation and the prediction of the counterfactual samples with the factual sample. To achieve the target, we devise three key building blocks for  retrospection, which successively perform \textit{representation comparison}, \textit{prediction comparison}, and \textit{fusion} . In particular, the module first compares the representation of each counterfactual sample with the factual sample; then compares their predictions accordingly; and fuses the comparison across the counterfactual samples.

\paragraph{Representation comparison.}
Given a pair of counterfactual sample $\bm{x}^*$ and factual sample $\bm{x}$, we believe the signals meaningful for making final decision lie in the difference of the samples and how the difference affects the classification. To distill such signals, we devise the representation comparison block as $\bm{y}_{\Delta} = f(\bm{x} - \bm{x}^* | \hat{\bm{\theta}})$,
where $\bm{y}_{\Delta} \in \mathcal{R}^C$ denotes the prediction of the representation difference $\bm{x} - \bm{x}^*$ given by the trained classification model.
Note that we leverage the trained model to enlighten how the content difference affects the classification since the model is trained to capture the connection between the textual patterns and the classes. 
It should be noted that we use a duplicate of the trained classification model for the representation comparison. That is to say, the training of the retrospection module will not affect the classification model.

\paragraph{Prediction comparison.} To retrospect the prediction $f(\bm{x} | \hat{\bm{\theta}})$, we devise a prediction comparison block to compare the predictions of each counterfactual and factual sample pair and distill patterns from $f(\bm{x} | \hat{\bm{\theta}})$, $f(\bm{x}^* | \hat{\bm{\theta}})$, and $\bm{y}_{\Delta}$. Inspired by the success of convolutional neural network (CNN) in capture local-region patterns, the block is devised as a CNN, which is formulated as:
\begin{equation}
{\small
\begin{aligned}\label{eq:hp}
    \bm{y}^* = 
    \text{CNN}\left(
        f(\bm{x} | \hat{\bm{\theta}}), f(\bm{x}^* | \hat{\bm{\theta}}), \bm{y}_{\Delta}
    \right),
\end{aligned}
}%
\end{equation}
where $\bm{y}^*$ denotes the retrospected prediction when comparing to $\bm{x}^*$. In particular, a \textit{stack layer} first stacks the three predictions as a matrix, which serves as an ``image'' to facilitate ``observing'' patterns. Formally,
$\bm{Y} = \left[f(\bm{x} | \hat{\bm{\theta}}), f(\bm{x}^* | \hat{\bm{\theta}}), \bm{y}_{\Delta}\right]$
where $\bm{Y} \in \mathbb{R}^{C \times 3}$. $\bm{Y}$ is then fed into an 1D convolution layer to capture the intra-class patterns across the predictions, which is formulated as:
\begin{equation}
{\small
\begin{aligned}\label{eq:hpcnn}
    \bm{H} = \sigma(\bm{Y} * \bm{F}),~\bm{H}_{ij} = \sigma( \bm{Y}_{:i} \bm{F}_j),
\end{aligned}
}%
\end{equation}
where $\bm{F} \in \mathbb{R}^{3 \times K}$ denotes the filters in the convolution layer, and $\sigma(\cdot)$ is an activation function such as \textit{GELU}~\cite{hendrycks2016gaussian}. $\bm{Y}_{:i}$ and $\bm{F}_j$ represent the $i$-th row of $\bm{Y}$ and the $j$-th column of $\bm{F}$, respectively. The filter $\bm{F}_j$ can learn rules for conducting retrospection. For instance, a filter $[1, -1, 0]$ means deducting the prediction of the counterfactual sample from that of the factual sample. The output $\bm{H} \in \mathbb{R}^{C \times K}$ is then flattened as a vector and fed into a fully-connected (FC) layer to capture the inter-class patterns. Formally,
\begin{equation}\label{eq:hpfc}\small
    \bm{y}^* = \bm{W} flatten(\bm{H}) + \bm{b},
\end{equation}
where $\bm{W}$ and $\bm{b}$ are model parameters.

\paragraph{Fusion.} The target is to fuse the retrospected predictions $\{\bm{y}^*\}$ into a final decision $\bm{y}$. Inspired by the success of pooling function in reading out patterns, we devise the block as $\bm{y} = pooling(\{\bm{y}^*\})$. As the fusion is performed after the pairwise comparison, we term it as \textit{late fusion}.

\paragraph{Training.} We update the parameters of the retrospection module by minimizing the classification loss over the factual training samples, which is:
\begin{equation}
{\small
\begin{aligned}\label{eq:obj-retrospection}
    \hat{\bm{\eta}} = \min_{\bm{\eta}} 
    & \sum_{\left({\bm{x}}, {{y}}\right) \in \mathcal{T}}
    l({y}, \bm{y}) + \lambda \|\bm{\eta}\|.
\end{aligned}
}%
\end{equation}
where $\lambda$ denotes the hyper-parameter to adjust the weight of the regularization term.

It should be noted that no existing research has uncovered the specific mechanism of retrospection in our brain, \ie the order of comparison and fusion is unclear. As such, 
we further devise two fusion strategies: \textit{middle fusion} and \textit{early fusion}, which performs fusion within the CNN, \ie during comparison, and before the CNN, respectively.

\noindent $\bullet$ \textit{Middle fusion} performs aggregation between the convolution layer and the FC layer. This fusion first calculates the latent comparison signals $\bm{H}$ for each pair of counterfactual and factual samples according to Equation~\ref{eq:hpcnn}. The aggregated signals $pooling(\{\bm{H}\})$ are then fed into the FC layer (Equation~\ref{eq:hpfc}) to obtain the final decision $\bm{y}$. 

\noindent $\bullet$ \textit{Early fusion} aggregates the counterfactual samples before performing comparison, which is formulated as $\tilde{\bm{x}}^* = pooling(\{\bm{x}^*\})$. In this way, the retrospection module is formulated as: $\bm{y} = \text{CNN}\left(f(\bm{x} | \hat{\bm{\theta}}), f(\tilde{\bm{x}}^* | \hat{\bm{\theta}}), f(\tilde{\bm{x}}^* - \bm{x} | \hat{\bm{\theta}}) \right).$
For all the three fusion methods, we can use either regular pooling function without parameter or parameterized pooling function~\cite{ying2018hierarchical} to enhance the expressiveness of the retrospection module. In our experiments, using a simple mean pooling achieves a performance that is comparable to the parameterized one in most cases (\cf Table~\ref{tab:fusion}).

\subsubsection{Generation Module}
The target is to construct counterfactual samples that are informative for retrospecting the decision on the target factual sample $\bm{x}$. As the task involves making decision among 
$C$ candidate classes, we believe that the key to generate representative counterfactual samples lies in imagining \textit{``what would the content be if the sample belongs to class $c$''}, \ie generating $C$ counterfactual samples $\{\bm{x}^*_c\}$. With the $C$ classes as the targets, the searching space of samples can also be largely narrowed down. Toward this end, we devise the generation module with two main considerations: 1) decomposing the factual sample $\bm{x}$ to distill contents irrelevant to the label of the sample $\bm{u} = d(\bm{x} | \bm{\omega})$; 2) injecting class $c$ into $\bm{u}$ to form the counterfactual sample $\bm{x}^*_c$.

\paragraph{Decomposition.} To distill $\bm{u}$, we need to recognize the connection between the content of the factual sample and each class. We thus account for class representations in the decomposition function. To align the sample space of the generation module with the retrospection module $h(\cdot)$ and the classification model $f(\cdot)$, we extract the parameters from the prediction layer of the trained classification model as the class representations. In particular, we extract the mapping matrix $\bm{W} \in \mathbb{R}^{C \times D}$ where the $c$-th row corresponds to class $c$. Note that we assume that the prediction layer has the same dimensionality as the latent representation, which is a common setting in most cutting edge language understanding models. The decomposition function is devised as a CNN to capture both the intra-dimension and inter-dimension connections between the factual sample and the classes.

\noindent $\bullet$ \textit{Stack layer.} The stack layer stacks the factual sample, class representations, and the element-wise product between sample and each class, which is formulated as:
$\bm{X} = [\bm{x}, \bm{W}^T, \bm{x} \odot \bm{W}^T]$.
$\bm{x} \odot \bm{W}^T \in \mathbb{R}^{D \times C}$ shed lights on how closely each dimension of $\bm{x}$ connect to each class, where large absolute value indicates closer connections.
    
\noindent $\bullet$ \textit{Convolution layer.} This layer uses 1D horizontal filters to learn patterns of deducting class relevant contents from the factual sample, which is formulated as $\bm{h} = pooling(\sigma(\bm{X} * \bm{F}^g))$.
$\bm{F}^g \in \mathbb{R}^{(2C + 1) \times L}$ denotes the filters where $L$ is the total number of filters. The output $\bm{h} \in \mathbb{R}^D$ is a hidden representation.
    
\noindent $\bullet$ \textit{FC layers.} We use two FC layers to capture the inter-dimension connections. Formally, $\bm{u} = \bm{W}^2 \sigma(\bm{W}^1 \bm{h} + \bm{b}^1) + \bm{b}^2$,
where $\bm{W}^2 \in \mathcal{R}^{D \times M}$, $\bm{W}^1 \in \mathcal{R}^{M \times D}$, $\bm{b}^2 \in \mathcal{R}^{D}$, and $\bm{b}^1 \in \mathcal{R}^{M}$ are learnable parameters. $M$ is a hyper-parameter to adjust the complexity of the decomposition function. Note that we can stack more layers to enhance the expressiveness of the function, whereas using two layers according to the universal approximation theorem~\cite{hornik1991approximation}.

We learn the parameters of the decomposition function from the counterfactual training samples by optimizing the following objective:
\begin{equation}
{\small
\begin{aligned}\label{eq:obj-generator}
    \min_{\bm{\omega}} \sum_{\left({\bm{x}}^*_c, c\right) \in \mathcal{T}^*} 
    &r\big(
        {\bm{u}}^*_c, \tilde{\bm{u}}_c
    \big) 
    + 
    \gamma 
    l\big(
        c, f({\bm{x}}^*_c - \bm{u}^*_c | \hat{\bm{\theta}})
    \big) \\
    &
    +
    r\big(
        {\bm{u}}, \tilde{\bm{u}}_c
    \big)
    +
    \gamma 
    l\big(
        {y}, f({\bm{x}} - {\bm{u}} | \hat{\bm{\theta}})
    \big),
\end{aligned}
}%
\end{equation}
where ${\bm{u}}^*_c = d({\bm{x}}^*_c | \bm{\omega})$ and ${\bm{u}} = d({\bm{x}} | \bm{\omega})$ are the decomposition results of the counterfactual sample ${\bm{x}}^*_c$ and the corresponding factual sample ${\bm{x}}$; $\tilde{\bm{u}}_c = \frac{1}{2}({\bm{x}} + {\bm{x}}^*_c)$ denotes the target value of the decomposition. The two terms $r(\cdot)$ and $l(\cdot)$ are Euclidean distance~\cite{dattorro2010convex} and classification loss. By minimizing the two terms, we encourage the decomposition result: 1) to be close to the target value $\tilde{\bm{u}}_c$; and 2) if being deducted from the original sample (\eg, ${\bm{x}} - {\bm{u}}$), the classification cannot be influenced. $\gamma$ is a hyper-parameter to balance the two terms.

The rationality of setting $\tilde{\bm{u}}_c = \frac{1}{2}({\bm{x}} + {\bm{x}}^*_c)$ as the target class irrelevant content of ${\bm{x}}$ and ${\bm{x}}^*_c$ comes from the parallelogram law~\cite{nash2003generalized}. Note that this pair of samples belong to two different classes where a decision boundary (a hyperplane) lies between the two classes ${y}$ and $c$. Considering that the sample ${\bm{x}}$ corresponds to a vector in the hidden space, we can decompose the vector into two components that are orthogonal and parallel to the decision boundary, \ie ${\bm{x}}^*_c = {\bm{o}}^*_c + {\bm{p}}^*_c$ and ${\bm{x}} = {\bm{o}} + {\bm{p}}$. Since the two samples belong to different classes, their orthogonal components are in opposite directions and their addition will only retain the parallel components, which are irrelevant to judging the class between ${y}$ and $c$\footnote{Note that we normalize all samples to be unit vectors in the decomposition function. Moreover, inspired by~\cite{parascandolo2018learning}, we train a decomposition function for each class, \ie class-specific parameters $\hat{\bm{\omega}}_c$}.

\paragraph{Injection.} Accordingly, given a testing sample $\bm{x}$, we can inject the orthogonal components towards class $c$ via $\bm{x}^*_c = 2 * d(\bm{x} | \hat{\bm{\omega}}_c) - \bm{x}$, which is the imagined content of the sample if it belongs to class $c$. In this way, for each testing sample, we conduct the injection over all the classes and construct $C$ counterfactual samples $\{\bm{x}^*_c\}$, which are then used in the retrospection module\footnote{The generation module consists of $C$ decomposition functions $d(\bm{x} | \hat{\bm{\omega}}_c)$ and the non-parametric injection function.}.

\subsection{Learning Paradigm with CRM}\label{ssec:paradigm}
The existing work~\cite{kaushik2019cfdata,zeng2020counterfactual} for language understanding typically follows the standard learning paradigm, \ie training a classification model over labeled data. Applying the proposed CRM indeed forms a new learning paradigm for constructing language understanding solutions. Algorithm~\ref{algo:paradigm-crm} illustrates the procedure of the new paradigm.


 
 
 
 

\begin{algorithm}[h]\small
	\caption{Learning paradigm with CRM}  
	\label{algo:paradigm-crm}
	\begin{algorithmic}[1]
		\Require Training data $\mathcal{T}$, $\mathcal{T}^*$. 
		
	    /* Training */
	    
	    \State Optimize Equation~\ref{eq:obj-model};  \algorithmiccomment{Classification model training}
	    
	    \State Optimize Equation~\ref{eq:obj-generator}; \algorithmiccomment{Generation module training}
	    
	    \State Optimize Equation~\ref{eq:obj-retrospection}; \algorithmiccomment{Retrospection module training}
	    
	    \State Return $\hat{\bm{\theta}}$, ${\hat{\bm{\omega}}_c}$, and $\hat{\bm{\eta}}$.

		/* Testing */
		\State Calculate $f(\bm{x} | \hat{\bm{\theta}})$; \algorithmiccomment{Classification model inference}
		
		\For{$c = 1 \to C$}
		\State $\bm{x}^*_c = 2 * g(\bm{x} | \hat{\bm{\omega}}_c) - \bm{x}$; \algorithmiccomment{Generation}
		\EndFor
		
		\State Calculate $h(\bm{x}, \{\bm{x}^*_c\}|\hat{\bm{\eta}}, \hat{\bm{\theta}})$; \algorithmiccomment{Retrospection}
	\end{algorithmic}
\end{algorithm}
\setlength{\textfloatsep}{0.1cm}
\section{Experiments}\label{sec:exp}
We conduct experiments on two representative language understanding tasks, SA and NLI, to answer the following research questions:

\noindent $\bullet$ \textbf{RQ1:} To what extent counterfacutal reasoning improves language understanding?

\noindent $\bullet$  \textbf{RQ2:} How does the design of the retrospection module affect the proposed CRM?

\noindent $\bullet$  \textbf{RQ3:} How effective are the counterfactual samples generated by the proposed generation module?

\subsection{Experiment Settings}\label{ssec:setting}

\textbf{Datasets.} We adopt the same datasets in~\cite{kaushik2019cfdata} for both tasks. The SA data are reviews from IMDb, which are labeled as either positive or negative. For each factual review, the dataset contains a manually constructed counterfactual sample where the crowd workers are asked to manipulate the text to reverse the label with the constraint of no gratuitous change. NLI is a three-way classification task with two sentences as inputs and the target of detecting their relation within entailment, contradiction, and neutral. For each factual sample, four counterfactual samples are given, which are constructed by editing either the first or the second sentence with target relations different to the label of the factual sample. 

\textbf{Classification models.} Owing to the extraordinary representational capacity of language model, fine-tuning pre-trained language model has become the emergent technique for solving language understanding tasks~\cite{jacob2019bert}. We select the widely used RoBERTa-base\footnote{\url{https://huggingface.co/roberta-base}.} and RoBERTa-large\footnote{\url{https://huggingface.co/roberta-large}.} for the consideration of the robustness of the RoBERTa~\cite{liu2019roberta} and our limited computation resources. For SA, we also test the classical Multi-Layer Perceptron (MLP)~\cite{teney2020+gs} with \textit{tf-idf} text features~\cite{schutze2008introduction} as inputs. For NLI, we further test RoBERTa-large-nli\footnote{\url{https://huggingface.co/roberta-large-mnli}.}, which has been fine-tuned on the large-scale MultiNLI dataset~\cite{williams2018broad}.

\textbf{Baselines.} As the proposed CRM leverages counterfactual samples, we compare CRM with three representative methods using counterfactual samples in language understanding tasks: 1) \textbf{+CF}~\cite{kaushik2019cfdata}, which uses counterfactual samples as data augmentation for model training; 2) \textbf{+GS}~\cite{teney2020+gs}, which compares the factual and counterfactual samples in model training through regularizing their gradients; and 3) \textbf{+CL}~\cite{liang2020col}, which compares the factual and counterfactual samples through a contrastive loss. Moreover, we report the performance of the testing model under \textit{Normal Training}, \ie training over factual samples only.

\textbf{Implementation.} We implement the proposed CRM with PyTorch 1.7.0 based on Hugging Face Transformer\footnote{\url{https://github.com/huggingface/transformers}.}, which is released at: \url{https://github.com/fulifeng/Counterfactual_Reasoning_Model}. In all cases, we follow the setting of +CF for training the classification model, which is a standard fine-tuning in~\cite{liu2019roberta}. We then use \textit{adam}~\cite{kingma2014adam} with learning rate of 0.001 to optimize the retrospection module and the generation module. For the retrospection module, we set the number of filters in the convolution layer $K$ as $10$, the weight for regularization $\lambda$ as $0$. As to the generation module, we set the number of convolution filters as $10$, the size of the hidden layer $M$ as $256$, and the weight for balancing Euclidean distance and classification loss $\gamma$ as $15$. We report the average classification accuracy over 5 different runs. For each repeat, we train the model with 20 epochs and select the model with the best performance on the validation set. 

\begin{table*}[tbhp]
\centering
    \begin{tabular}{l|c|ccc|c|c}
    \hline
    \multicolumn{7}{c}{\cellcolor[HTML]{C0C0C0}Sentiment Classification} \\ \hline
    Backbone & Normal Training & +CF & +GS & +CL & +CRM & \textit{RI} \\ \hline
    MLP & 86.9$\pm$0.5 & 85.3$\pm$0.3 & 84.6$\pm$0.4 & - & \textbf{98.6$\pm$0.2} & 15.6\% \\ 
    RoBERTa-base & 93.2$\pm$0.6  & 92.3$\pm$0.7 & 92.2$\pm$0.9 & 91.8$\pm$1.1 & \textbf{97.5$\pm$0.3} & 5.7\% \\ 
    RoBERTa-large & 93.6$\pm$0.6 & 93.4$\pm$0.4 & 93.1$\pm$0.5 & 94.1$\pm$0.4 & \textbf{98.2$\pm$0.3} & 5.1\% \\ \hline
    \multicolumn{7}{c}{\cellcolor[HTML]{C0C0C0}Natural Language Inference} \\ \hline
    Backbone & Normal Training & +CF & +GS & +CL & +CRM & \textit{RI} \\ \hline
    RoBERTa-base & 83.5$\pm$0.8 & 83.4$\pm$0.9 & 83.8$\pm$1.7 & 84.1$\pm$1.1 & \textbf{91.5$\pm$1.6} & 9.7\% \\ 
    RoBERTa-large & 87.9$\pm$1.7 & 85.8$\pm$1.2 & 86.2$\pm$1.2 & 86.5$\pm$1.6 & \textbf{93.8$\pm$1.9} & 9.3\% \\ 
    RoBERTa-large-nli & 89.4$\pm$0.7 & 88.2$\pm$1.0 & 87.2$\pm$1.4 & 88.2$\pm$1.0 & \textbf{94.4$\pm$1.2} & 7.1\% \\ \hline
    \end{tabular}%
\vspace{-0.2cm}
\caption{Performance of the proposed CRM (Early Fusion) and baselines on the SA and NLI tasks. RI means the relative performance improvement achieved by +CRM over the classification model without CRM, \ie +CF.}
\vspace{-0.4cm}
\label{tab:perf_comp}
\end{table*}

\subsection{Performance Comparison (RQ1)}
We first use the handcrafted counterfactual samples to demonstrate the effectiveness of counterfactual reasoning in the inference stage of language understanding model, which can be seen as using a golden standard generation module to provide counterfactual samples for the retrospection module. Note that we do not use the label of counterfactual samples in the testing set. Table~\ref{tab:perf_comp} shows the performance of the compared methods on the two tasks. From the table, we observe that:
\begin{itemize}[parsep=0pt, leftmargin=*]
    \item +CRM largely outperforms all the baseline methods in all cases. As compared to +CF, the same classification model without CRM in the testing phase, +CRM achieves relative performance improvement up to 15.6\%. The performance gain is attributed to the retrospection module, which justifies the rationality and effectiveness of incorporating counterfactual thinking into the inference stage of language understanding model. In other words, by comparing the factual sample with its counterfactual samples, the retrospection module indeed makes more accurate decisions.
    \item On the SA task, a huge gap (85.3 $\leftrightarrow$ 93.4) lies in the performance of the shallow model MLP and the deep RoBERTa-base/RoBERTa-large. When applying +CRM, MLP achieves a performance that is comparable to the deep models. The result indicates that counterfactual reasoning can compensate for the disadvantages caused by the insufficient model representational capacity. In addition, the result reflects that CRM brings cognitive ability beyond recognizing textual patterns. If the retrospection module only facilitates capturing the correlation between textual patterns and classes, such simple model cannot bridge the huge gap of representational capacity between MLP and RoBERTa-large.
    \item The performance of baseline methods are comparable to each other in most cases, \ie incorporating counterfactual samples into model training does not necessarily improve the testing performance on factual samples. This result is consistent with~\cite{kaushik2019cfdata}, which is reasonable since these methods are devised for enhancing the generalization ability, especially for the out-of-distribution testing samples, which can sacrifice the performance on normal testing samples. Besides, the result indicates that training with counterfactual samples is insufficient for achieving counterfactual thinking, which reflects the rationality of enhancing the inference paradigm with a decision making procedure.
\end{itemize}

\begin{figure}[h]
	\centering
	\vspace{-0.3cm}
	\mbox{
	    \hspace{-0.15in}
		\subfigure[Sentiment analysis]{
			\label{fig:conf-comp-sa}
			\includegraphics[width=0.24\textwidth]{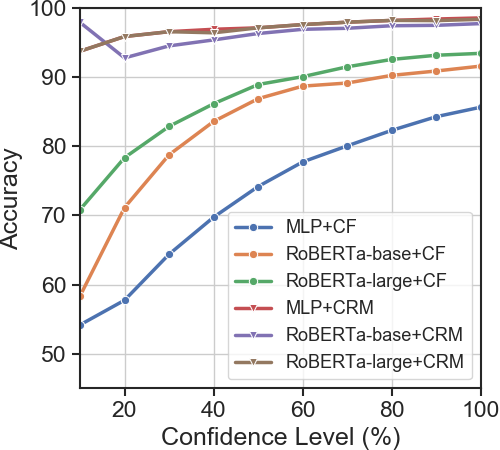}
		}
		\hspace{-0.15in}
		\subfigure[Natural language inference]{
			\label{fig:conf-comp-nli}
			\includegraphics[width=0.24\textwidth]{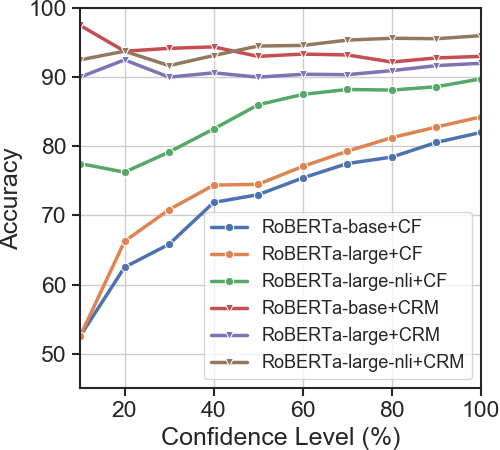}
		}
	}
	\vspace{-0.4cm}
	\caption{Prediction performance of +CF and +CRM over testing samples at different confidence levels.
	}
	\vspace{-0.1cm}
	\label{fig:conf-comp}
\end{figure}
\paragraph{Performance on hard samples.} Furthermore, we investigate whether the proposed CRM facilitate dealing with hard samples. Recall that we split the testing samples into 10 groups according to the confidence of the classification model, \ie +CF (\cf Section~\ref{sec:pilot}). We perform group-wise comparison between +CF and +CRM. Figure~\ref{fig:conf-comp} shows the performance of all the classification models with +CF and +CRM. From the figures, 1) we observe that the performance of +CRM is stable across different confidence levels, whereas the performance of the classification model shows a clear decreasing trend as the confidence level decreases from 10 to 1. The result indicates that the retrospection module is insensitive to the confidence of the classification model. 2) In all cases, +CRM achieves the largest performance gain at the first group with confidence level of 1, \ie the hardest group to the classification model. For instance, the improvement reaches 85.7\% on the RoBERTa-base model for the NLI task. The large improvements further justifies the effectiveness of the retrospection module, \ie comparing the prediction of factual samples to counterfactual samples indeed facilitates dealing with hard samples.

\begin{table}[htbp]
\centering
    \begin{tabular}{l|c|c}
    \hline
    \multicolumn{3}{c}{\cellcolor[HTML]{C0C0C0}Sentiment Classification} \\ \hline
    Backbone & Implicit & +CRM \\ \hline
    MLP & 79.3$\pm$0.2 & \textbf{98.6$\pm$0.2} \\ 
    RoBERTa-base & 94.7$\pm$0.6 & \textbf{97.5$\pm$0.3} \\ 
    RoBERTa-large & 98.0$\pm$0.4 & 98.2$\pm$0.3 \\ \hline
    \multicolumn{3}{c}{\cellcolor[HTML]{C0C0C0}Natural Language Inference} \\ \hline
    Backbone & Implicit & +CRM \\ \hline
    RoBERTa-base & 81.9$\pm$3.5 & \textbf{91.5$\pm$1.6} \\ 
    RoBERTa-large & 87.4$\pm$2.2 & \textbf{93.8$\pm$1.9} \\ 
    RoBERTa-large-nli & 88.8$\pm$1.6 & \textbf{94.4$\pm$1.2} \\ \hline
    \end{tabular}%
\vspace{-0.2cm}
\caption{Performance comparison of implicit modeling (end-to-end model) and explicit modeling (CRM) of counterfactual thinking.}
\vspace{-0.4cm}
\label{tab:implicit_VS_CRM}
\end{table}
\paragraph{CRM \textit{V.S.} implicit modeling.} According to the uniform approximation theorem~\cite{hornik1991approximation}, the CRM can also be approximated by a deep neural network. We thus investigate whether counterfactual thinking can be learned in an implicit manner. In particular, we evaluate a model that takes both the factual sample and counterfactual samples as inputs to make prediction for the factual one. Table~\ref{tab:implicit_VS_CRM} shows the performance, where we have the following observations: 1) The implicit modeling performs much worse than the proposed CRM in most cases, which justifies the effectiveness of the retrospection module and the rationality of modeling comparison explicitly. 2) On the NLI task, RoBERTa-base+CRM outperforms RoBERTa-large (implicit), which means that the superior performance of CRM is not because of the additional model parameters introduced by the retrospection module, but the explicit comparison between factual and counterfactual samples. 

\subsection{In-depth Analysis}
\begin{table*}[]
\centering
\begin{tabular}{l|c|cc|cc|cc}
\hline
Backbone           & +CF & EF & \textit{RI} & LF & \textit{RI} & MF & \textit{RI} \\ \hline \hline
RoBERTa-base       & 83.4$\pm$0.9 & 91.5$\pm$1.6 & 9.7\% &      \textbf{92.8$\pm$1.8} & 11.3\% &         89.6$\pm$2.0 & 7.4\% \\
RoBERTa-large      & 85.8$\pm$1.2 & 93.8$\pm$1.9 & 9.3\% &      \textbf{95.3$\pm$0.7} & 11.1\% &     93.4$\pm$1.7 & 8.9\% \\
RoBERTa-large-nli & 88.2$\pm$1.0 & 94.4$\pm$1.2 & 7.1\% &      93.8$\pm$0.4 & 6.4\% &      \textbf{94.7$\pm$1.3} & 7.4\% \\ \hline
\end{tabular}
\vspace{-0.2cm}
\caption{Performance of the proposed CRM based on early fusion (EF), late fusion (LF), or middle fusion (MF) on the NLI task. RI represents the relative performance improvement over the +CF method.}
\vspace{-0.2cm}
\label{tab:fusion}
\end{table*}
\paragraph{Effects of retrospection module design (RQ2).}
Note that the order of comparison and fusion in the retrospection mechanism of us humans is still unclear. We investigate how the fusion strategies influence the effectiveness of the proposed CRM. Table~\ref{tab:fusion} shows the performance of CRM based on the early fusion (EF), late fusion (LF), and middle fusion (MF) on the NLI task. We omit the comparison on the SA task since the dataset only has one counterfactual sample for the target factual sample. For both EF and LF, we use the mean pooling as the pooling function. As to MF, we use a pooling function that is equipped with self-attention~\cite{vaswani2017attention}. The reasons of this setting are twofold: 1) using mean pooling will make LF and MF equivalent since the FC layer in the retrospection module is a linear mapping. Note that LF performs pooling after the FC layer, while the pooling function of MF is just before the FC layer. 2) The comparison between the LF and MF can thus shed light on whether parameterized pooling function can benefit the retrospection.

From the table, we can observe that, in most cases, CRM based on different fusion strategies achieve performance comparable to each other. It indicates that the retrospection is insensitive to the order of fusion and the comparison between counterfactual and factual samples. Considering that MF with mean pooling is equivalent to LF, we can see that the benefit of parameterized pooling function is limited. In particular, MF only performs better than LF on one of the three testing models. 

\paragraph{Effects of generation module (RQ3).}
We then investigate whether the proposed generation module constructs useful counterfactual samples for retrospection. We train and test the retrospection module (using EF) with the generated samples on RoBERTa-large on the SA task. We omit the experiments of other settings for saving computation resources. In this way, the model achieves an accuracy of 94.5 which is better than +CF (93.4) but worse than +CRM with manually constructed counterfactual samples (98.2) (\cf Table~\ref{tab:perf_comp}). The result indicates that the generated samples indeed facilitate the retrospection while the generation quality can be further improved. Moreover, on the testing samples at confidence level of 1, using the generated samples achieves an accuracy of 81.3 which is much better than +CF (70.8) (\cf Figure~\ref{fig:conf-comp}). The generated samples indeed benefit the decision making over hard testing samples.
\section{Related Work}
\textbf{Counterfactual sample.} Constructing counterfactual samples has become an emergent data augmentation technique in natural language processing, which has been used in a wide spectral of language understanding tasks, including SA~\cite{kaushik2019cfdata,yang2020generating}, NLI~\cite{kaushik2019cfdata}, named entity recognition~\cite{zeng2020counterfactual} question answering~\cite{chen2020counterfactual}, dialogue system~\cite{zhu2020counterfactual}, vision-language navigation~\cite{fu2020counterfactual}. Beyond data augmentation under the standard supervised learning paradigm, a line of research explores to incorporate counterfactual samples into other learning paradigms such as
adversarial training~\cite{zhu2020counterfactual,fu2020counterfactual,teney2020+gs} and contrastive learning~\cite{liang2020col}. This work lies in an orthogonal direction that incorporates counterfactual samples into the decision making procedure of model inference.

\textbf{Counterfactual inference.} A line of research attempts to enable deep neural networks with counterfactual thinking by incorporating counterfactual inference~\cite{yue2021counterfactual,wang2020click,niu2020counterfactual,tang2020long,feng2020should}. These methods perform counterfactual inference over the model predictions according to a pre-defined causal graph. Due to the requirement of causal graph, such methods are hard to be generalized to different tasks. Our method does not suffer from such limitation since working on the counterfactual samples which can be generated without a comprehensive causal graph.

\textbf{Hard sample.} A wide spectral of machine learning techniques are related to dealing with the hard samples in language understanding. For instance, adversarial training~\cite{khashabi2020more} enhances the model robustness against perturbations and attacks, which are hard samples for normally trained models. Debiased training~\cite{tu2020empirical,utama2020mind} eliminates the spurious correlation or bias in training data to enhance the generalization ability and deal with out-of-distribution samples. In addition to the training phase, a few inference techniques might improve the model performance on hard samples, including posterior regularization~\cite{srivastava2018zero} and causal inference~\cite{yu2020counterfactual,niu2020counterfactual}. However, both techniques require domain knowledge such as prior or causal graph tailored for specific applications. On the contrary, this work provides a general paradigm that can be used for most language understanding tasks.
\section{Conclusion}
In this work, we pointed out the issue of standard inference of existing language understanding models. We proposed a Counterfactual Reasoning Model which empowers the trained model with a high-level cognitive ability, counterfactual thinking. By applying the proposed CRM, we formed a new paradigm for building language understanding solutions. We conducted extensive experiments, which validate the effectiveness of our proposal, especially in dealing with hard samples. 

This work opens up a new research direction about the decision making procedure in testing phase. In the future, we will explore sequential decision procedure to resolve the constraint on the number of constructed counterfactual samples. In addition, we will investigate generation module for language understanding with unsupervised generative techniques~\cite{sauer2021counterfactual}.

\section*{Acknowledgments}
This research is supported by the Sea-NExT Joint Lab, Singapore MOE AcRF T2, National Natural Science Foundation of China (U19A2079) and National Key Research and Development Program of China (2020AAA0106000). Our thanks also go to all the anonymous reviewers for their valuable suggestions. 

\bibliographystyle{acl_natbib}
\bibliography{references}

\end{document}